\documentclass{article}
\usepackage{arxiv}

\newtheorem{example}{Example}
\newtheorem{definition}{Definition}
\usepackage{cite}
\usepackage{amsmath,amssymb,amsfonts}
\usepackage{graphicx}
\usepackage{textcomp}
\usepackage{xcolor}

\AtBeginDocument{%
  \providecommand\BibTeX{{\rm B\kern-.05em{\sc i\kern-.025em b}\kern-.08em
    T\kern-.1667em\lower.7ex\hbox{E}\kern-.125emX}}}

\makeatother

\usepackage{amsfonts}
\usepackage[multiple]{footmisc}
\usepackage{booktabs}
\usepackage{dsfont}
\usepackage[utf8]{inputenc}
\usepackage{balance}
\usepackage{siunitx}
\usepackage{multirow}
\usepackage{graphicx}
\usepackage{listings}
\usepackage{commath}
\usepackage{hyperref}
\usepackage{color}
\usepackage{url}
\usepackage{caption}
\usepackage{alltt}
\usepackage{mathrsfs}
\usepackage{float}
\usepackage{subcaption}
\usepackage{graphicx,fancyvrb}
\usepackage[linesnumbered,ruled,vlined]{algorithm2e}
\usepackage{algpseudocode}
\usepackage{amsmath}
\usepackage{booktabs}

\usepackage{mathtools}
\usepackage{caption}
\usepackage{float}
\usepackage{capt-of}
\usepackage{booktabs}
\usepackage{colortbl}
\usepackage{xcolor}
\usepackage{xfrac}
\usepackage{footnote}

 \usepackage{enumitem}
\usepackage{array}
\usepackage{arydshln}

\usepackage{cleveref}

\definecolor{mygreen}{rgb}{0,0.6,0}
\definecolor{myred}{rgb}{0.6,0,0}
\definecolor{mygray}{rgb}{0.5,0.5,0.5}
\definecolor{mymauve}{rgb}{0.58,0,0.82}
\definecolor{myblue}{rgb}{0,0,1}

\usepackage{listings}
\lstnewenvironment{VerbatimText}[1][]{
    
    \lstset{fancyvrb=true,basicstyle=\footnotesize,captionpos=b,xleftmargin=2em,#1}
}{}

\usepackage{booktabs}
\usepackage{pgfplots}
\pgfplotsset{compat=1.7}
\usepackage{pgfplotstable}

\PassOptionsToPackage{hyphens}{url}\usepackage{hyperref}

\usepackage{amsmath}

\newcommand{\cm}{{{ M }}}

\newcommand{\Exog}{{{ \mb U }}}

\newcommand{\Rse}{\mc{F}}

\newcommand{\Pa}{\mb{Pa}}
\newcommand{\pa}{\mb{pa}}

\newcommand{\pr}{{\tt \mathrm{Pr}}}
\newcommand{\sys}{\textsc{FairDebugger}\xspace}

\newcommand{\mc}[1]{\mathcal{#1}}

\newcommand{\ignore}[1]{}

\definecolor{black}{rgb}{0,0,0}
\definecolor{grey}{rgb}{0.8,0.8,0.8}
\definecolor{red}{rgb}{1,0,0}
\definecolor{green}{rgb}{0,1,0}
\definecolor{darkgreen}{rgb}{0,0.5,0}
\definecolor{darkpurple}{rgb}{0.5,0,0.5}
\definecolor{darkdarkpurple}{rgb}{0.3,0,0.3}
\definecolor{blue}{rgb}{0,0,1}
\definecolor{shadegreen}{rgb}{0.95,1,0.95}
\definecolor{shadeblue}{rgb}{0.95,0.95,1}
\definecolor{shadered}{rgb}{1,0.85,0.85}
\definecolor{shadegrey}{rgb}{0.85,0.85,0.85}
\definecolor{oddRowGrey}{rgb}{0.80,0.80,0.80}
\definecolor{evenRowGrey}{rgb}{0.85,0.85,0.85}
\definecolor{lightpurple}{rgb}{0.88,1.0,1.0}

\newcommand{\RNum}[1]{\uppercase\expandafter{\romannumeral #1\relax}}

\newcommand{\mb}[1]{{\mathbf{#1}}}
\newtheorem{defn}{Definition}[section]

\newtheorem{col}[defn]{Colollary}

\newcommand{\proj}[1]{{\Pi}}
\newcommand{\sel}[1]{{\sigma}}

\newcommand{\cut}[1]{}
\newcommand{\eat}[1]{}

\usepackage[multiple]{footmisc}

\SetKwInput{KwInput}{Input}
\SetKwInput{KwOutput}{Output}

\newcommand{\delsubset}{T}

\begin{document}

\title{Example-based Explanations for Random Forests using Machine Unlearning}
\author{
  Tanmay Surve\\
    Computer and Information Technology\\
   Purdue University\\
   \texttt{tsurve@purdue.edu} \\
   \And
 Romila Pradhan\\
Computer and Information Technology\\
   Purdue University\\
   \texttt{rpradhan@purdue.edu} \\
}
\maketitle

\begin{abstract}
    Tree-based machine learning models, such as decision trees and random forests, have been hugely successful in classification tasks primarily because of their predictive power in supervised learning tasks and ease of interpretation. Despite their popularity and power, these models have been found to produce unexpected or discriminatory outcomes. Given their overwhelming success for most tasks, it is of interest to identify sources of their unexpected and discriminatory behavior. However, there has not been much work on understanding and debugging tree-based classifiers in the context of fairness.
    
    We introduce \sys, a system that utilizes recent advances in \textit{machine unlearning} research to identify training data subsets responsible for instances of fairness violations in the outcomes of a random forest classifier. \sys generates top-$k$ explanations (in the form of \textit{coherent} training data subsets) for model unfairness. Toward this goal, \sys first utilizes machine unlearning to estimate the change in the tree structures of the random forest when parts of the underlying training data are removed, and then leverages the apriori algorithm from frequent itemset mining to reduce the subset search space.   We empirically evaluate our approach on three real-world datasets, and demonstrate that the explanations generated by \sys are consistent with insights from prior studies on these datasets.
\end{abstract}

\section{Introduction}
\label{sec:intro}
 Machine learning (ML) is fast becoming the standard choice for data science applications that involve automated decision-making in sensitive domains such as finance, healthcare, crime prevention, and justice management. Designed carefully, ML-based systems have the potential to eliminate the undesirable aspects of human decision-making such as biased judgments. However, concern continues to mount that these systems reinforce systemic biases and discrimination often reflected in their training data.
 Technology giants routinely come under the radar for discriminating against people based on their race, zip codes and perceived gender~\cite{fb-housing,amazonhire2018,10.1145/2702123.2702520}, self-driving cars are less accurate at detecting pedestrians with darker skin tones~\cite{self-driving-cars}. Such discriminatory outcomes are harmful not only because they violate human rights but also because they impede and undermine societal trust in machine learning.
 
The need to \textit{debug} and \textit{explain} the causes for unexpected and discriminatory model behavior has propelled advances in the field of Explainable Artificial Intelligence (XAI)~\cite{arrieta2020explainable,guidotti2018survey,molnar2020interpretable} that refers to the ability of an AI-based system to explain its decisions and actions in a way that is understandable, accountable, and transparent to humans.
However, much of XAI research has been centered on generating \textit{feature-based explanations} that explain the behavior of an ML model in terms of the input features or attributes of its training data~\cite{lundberg2017unified,ribeiro2016should,ribeiro2018anchors,galhotra2021explaining,mothilal2020explaining}. In contrast, with the recent explosion of \textit{data-centric} revolution in AI, recent research has focused on generating \textit{example-based explanations} that explain ML model behavior in terms of particular data instances that the model has been trained on~\cite{koh2017understanding,basu2020second,10.1145/3514221.3517886}. 
These efforts, however, have focused on a niche class of ML models--- parametric models that have a convex loss function which is also twice-differentiable. On the other hand, non-parametric models, such as tree-based models, are widely used primarily because of their predictive power in supervised learning tasks and ease of interpretation. Existing techniques for generating feature-based explanations are applicable to tree-based models, such as decision trees and tree ensembles (including random forests and gradient-boosted decision trees)~\cite{10.1023/A:1010933404324,friedman2001greedy}; however, the problem of generating example-based explanations for tree-based models has not been explored well. Note that since tree-based models are non-parametric, the aforementioned approaches that focus on parametric models with a twice-differentiable convex loss function are not directly applicable to tree-based models. Given their overwhelming success for most prediction tasks, we are interested in generating example-based explanations for tree-based models and identifying \textit{root causes} for their unexpected and discriminatory behaviour. 

Discriminatory behavior, also termed \textit{fairness} in the algorithmic literature is broadly categorized as \textit{individual fairness}, \textit{group fairness} and \textit{causal fairness}. Individual fairness states~\cite{dwork2012fairness} that similar individuals must be treated similarly. Group fairness~\cite{verma2018fairness,10.1145/3457607} mandates parity between group of individuals belonging to different sensitive groups (e.g., males vs. non-males, Asians vs. non-Asians). Causal fairness, on the other hand, considers if features have a causal effect on fairness of outcomes~\cite{counterfactualfairness,chiappa2019path}. In this paper, we focus on notions of group fairness which we detail further in Section~\ref{sec:notations}.

\begin{figure*}[t]
    \centering
    \includegraphics[scale=0.55, bb=10 -10 900 400]{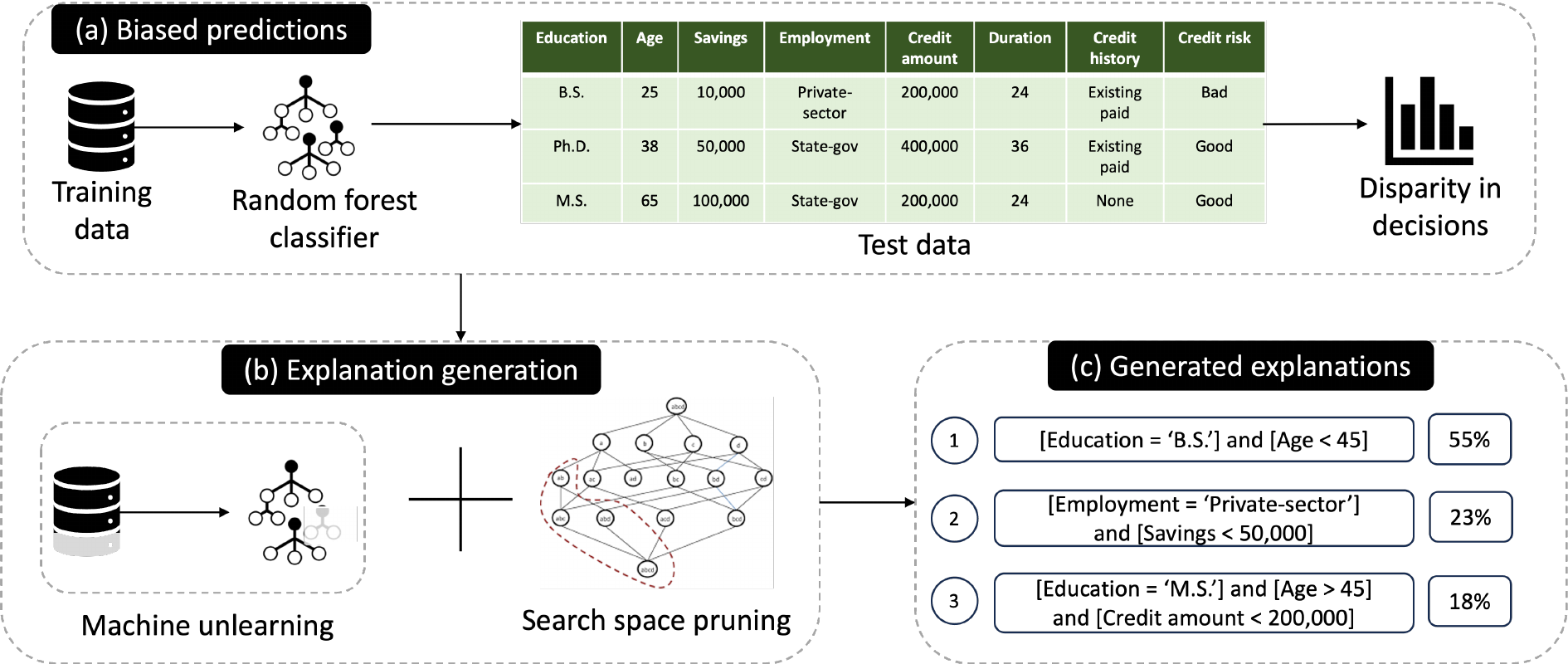}
    \caption{\textbf{An overview of \sys.} (a)~Given a random forest classifier trained on some training data, the classifier generates biased predictions on some test data. (b)~\sys uses machine unlearning and subset search space pruning techniques to determine the (c) top-$k$ training data subsets responsible for the biased predictions, along with the improvement in fairness of the updated model trained after deleting the susbet from the training data.}
    \label{fig:system}
    \vspace{-2mm}
\end{figure*}

Consider the following example (Example~\ref{ex:ex1}) that illustrates the need for generating example-based explanations for random forest classifiers.

\begin{example}
    [German Credit Dataset] Consider a data scientist Anne working on a classification pipeline that ingests demographic and financial information about individuals and learns a random forest classifier that is used to predict whether an individual is a good credit risk (and hence, should be granted a loan) or a bad credit risk (and hence, their loan application should be rejected). Anne observes that the learned model has a high accuracy ($\sim 91\%$) but also exhibits disparity in its predictions for individuals belonging to different age groups. In particular, she finds that older individuals (aged $45$ years and more) are $\sim 10\%$ more likely to be classified as good credit risks compared to younger individuals (aged $<45$ years). As a \textit{responsible} data scientist, Anne is concerned with this high disparity in predictions across age groups and wants to find out the reason for these biased decisions. Ideally, if she could learn what parts of the training data are causing the disparity, she might further inspect those parts for potential errors that might indicate issues in the earlier stages of the data science pipeline, and subsequently take remedial actions. For example, she might find out that some of the data instances were mislabeled and engage with a human annotator to correct those labels. She might also discover that there was not enough representation for individuals with particular demographic information and take remedial actions by acquiring additional data for that demographic group. Toward this goal, Anne tries out popular XAI techniques, such as SHAP~\cite{lundberg2017unified} and LIME~\cite{ribeiro2016should}, that generate feature-based explanations and and learns that individuals' age, their credit history and the status of their checking account play an important role for the model's predictions. Similar to Gopher~\cite{10.1145/3514221.3517886}, she wants to generate example-based explanations that identify coherent subsets of the data (e.g., (\textsf{Age $> 45$}) $\wedge$ (\textsf{Gender=Female})) that might be responsible for the disparity. However, Gopher utilizes the notion of influence functions~\cite{pmlr-v70-koh17a,basu2020second} which are not applicable to random forest classifiers because of their non-parametric nature. As a result, Anne is unable to obtain any insights into parts of the training data that might have caused this disparity and need further attention.
    \label{ex:ex1}
\end{example}
 
Generating example-based explanations is a computationally expensive task; to identify data instances that are responsible for biased decisions, their effect on the downstream model bias should be computed and the computation should be performed in an efficient manner. To judiciously utilize the data scientist, explanations should be presented in a format that is human-understandable and compact and should also quantify the contribution of instances to model bias. Assuming we have access to a data scientist or data curator who can inspect data instances for potential issues and errors, what are the top-k example-based explanations for bias in a random forest classifier that should be presented to them? 

The task of generating the top-k example-based explanations for a random forest classifier is challenging because of a number of reasons. First, we need to generate explanations that are compact and human-understandable. Second, for each such explanation, we need to compute the effect of the explanation on downstream model bias. To compute this effect, we need to quantify the contribution of the explanation to model bias without having to retrain the model. Finally, machine learning datasets typically have a large number of data instances and features, thus limiting the ability to compute the contribution to a small fraction of the entire data.

\noindent\textbf{Human-understandable explanations:} To tackle the first challenge, we generate explanations in the form of predicates that are conjunctions of literals $\wedge_{j} X_j$ $op$ $v_j$ where $X_j$ is a feature, $v_j$ is a corresponding feature value and $op$ can be one of $=, \neq, <, \le, \ge, >$. For instance, a possible explanation for the scenario in Example~\ref{ex:ex1} would be (\textsf{Age $> 45$}) $\wedge$ (\textsf{Gender=Female}). Such explanations represent subsets of the training data and are more informative  compared to individual data instances because they highlight potential issues in the earlier stages of the data science pipeline for specific subsets, and hence may reveal underlying discriminatory issues with how the data instances in these subsets are pre-processed.

\noindent\textbf{Responsible subset and contribution to bias:} We need a way to determine which subset of the training data instances cause the discriminatory outcome. We call such subsets \textit{responsible} subsets and quantify their contribution to model bias by computing the change in model bias when a new model is trained after removing the subset from the training data.

Identifying responsible subsets is challenging because of two reasons. First, to compute the contribution of a subset to model bias, we need to retrain the model without the subset in the training data. Retraining the model is computationally expensive; we need a way to estimate the contribution of subsets without retraining the model. Toward this goal, we utilize recent advances made in the upcoming field of machine unlearning~\cite{machine_unlearning_survey,bourtoule2019machine} that refers to the ability of a model to remove the effect of a few training data instances on model predictions. Second, for identifying subsets with the highest contribution to bias, we need to evaluate all possible predicates, which is exponential in the number of attributes and their cardinalities, and hence is infeasible in practice. To address this challenge, we utilize the apriori algorithm from frequent itemset mining~\cite{10.5555/645920.672836}.

\noindent \textbf{Summary of contributions.} In summary, our contributions are as follows: 
\begin{itemize}
    \item We present \sys, a system that facilitates end-to-end training data debugging for explaining bias in random forest classifiers. \sys identifies the subsets of the training data that are the most responsible for model bias. 
    \item We formalize the notion of contribution of subsets and present a system that searches for subsets with the highest contribution to model bias (Section~\ref{sec:notations})
    \item We leverage concepts from \textit{machine unlearning} to estimate the contribution of training data subsets to model bias (Section~\ref{sec:machine-unlearning}). To the best of our knowledge, \sys is the first framework that uses concepts from {machine unlearning} for the problem of fairness debugging. We present an algorithm that efficiently searches over training data subsets to generate top-$k$ explanations for bias in a random forest classifier (Section~\ref{sec:explanations}).
    \item We provide experimental evaluation on real-world datasets, and show that for random forest classifiers, the generated explanations are consistent with insights from prior studies (Section~\ref{sec:experiments}).

\end{itemize}

\section{Problem Definition}
\label{sec:notations}
In this section, we introduce the notations used throughout the paper, and then present relevant background information on classification and algorithmic fairness. We then formally introduce the problem we are solving in this paper.

\subsection{Preliminaries}
\noindent\textbf{Classification.} We consider the problem of binary classification and assume an instance space $\mathcal{X} \subseteq \mathbb{R}^p$ and binary labels $\mathcal{Y} = \{0,1\}$. Let $\mathcal{D} = \{(x_i, y_i)\}_{i=1}^n$ be a training dataset where each instance $x_i \in \mathcal{X}$ is a $p$-dimensional vector $(x_i,j )_{j=1}^p$
 and $y_i \in \mathcal{Y}$. The set of possible attributes is denoted by $\mathbf{X} = \{j\}_{j=1}^p$.
%
Let  $\mathcal{A} : \mathcal{D} \rightarrow \mathcal{H}$ represent a learning algorithm defined as a function from dataset $\mathcal{D}$ to a model in the hypothesis space $\mathcal{H}$. 
Let $h \in \mathcal{H}$ be the learned model obtained by training learning algorithm $\mathcal{A}$ on $\mathcal{D}$, and $\hat{Y}$ be the output space such that $\hat{y}=h(x)$ is its prediction on a test data instance $x \in \mathcal{X}$. 

\vspace{1pt}\noindent\textbf{Group fairness.} Given a binary classifier $h \in \mathcal{H}$ with output $\hat{Y}$ and a protected attribute $S \in \mathbf{X}$  (such as gender, race, age etc.), we interpret $\hat{Y}=1$ as a favorable (positive) prediction and $\hat{Y}=0$ as an unfavorable (negative) prediction. We assume the domain of $S$, \textsf{Dom}($S$) $= \{0, 1\}$ where $S = 1$ indicates a privileged and $S=0$ indicates a protected group (e.g.,
males and non-males, respectively). Group fairness  mandates that individuals belonging to different groups must be treated similarly. The notion of similarity in treatment is captured by different associative notions of fairness~\cite{verma2018fairness,10.1145/3457607,chouldechova2017fair}. We focus on the following widely used notions of group fairness:
\begin{itemize}[leftmargin=*]
    \item \textbf{Statistical parity:} A classifier $h$ satisfies statistical parity if both the protected and the privileged groups have the same probability of being predicted the positive outcome i.e., $P(\hat{Y} = 1|S = 0) = P(\hat{Y} = 1| S = 1)$.
    \item \textbf{Equalized odds:} A classifier $h$ satisfies equalized odds if the predictions $\hat{Y}$ and the sensitive attribute $S$ are independent conditional on the true labels $Y$, i.e., 
    $P(\hat{Y} = 1|S = 0, Y = y) = P(\hat{Y} = 1| S = 1, Y = y),$
    where $y \in \{0,1\}$.
    This definition states that the probability of an individual with a positive label being correctly assigned a positive outcome and the probability of an individual with a negative label being incorrectly assigned a negative outcome should be the same for both the protected and privileged group. In other words, the protected and privileged groups should have equal true positive rate and equal false positive rate.

    \item \textbf{Predictive parity:} A classifier $h$ satisfies predictive parity if $P({Y} = 1|S = 0, \hat{Y} = 1) = P({Y} = 1| S = 1, \hat{Y} = 1)$ i.e., the likelihood of a positive label among individuals predicted as having a positive outcome is the same regardless of group membership. 
\end{itemize}
These fairness metrics can be computed on both the training data predictions and the test data predictions. For a given dataset $\mathcal{D}$, fairness metric $\mathcal{F}: \mathcal{H} \times \mathcal{D} \rightarrow \mathbb{R}$ quantifies a given notion of group fairness computed over  $\mathcal{D}$. For example, for the notion of statistical parity, $\mathcal{F}(h, \mathcal{D}) = P(\hat{Y} = 1|S = 0) - P(\hat{Y} = 1| S = 1)$ where $P(\hat{Y} = 1 | S = s), s \in \{0, 1\}$ is estimated on $\mathcal{D}$ using the prediction probabilities obtained by applying classifier $h$ on $\mathcal{D}$.
To satisfy group fairness, we introduce the notion of \textit{group fairness violation} as follows:
\begin{definition}
    \textbf{Group fairness violation.} Given a dataset $\mathcal{D'}$, a random forest classifier $h$, and a fairness metric $\mathcal{F}$, group fairness violation occurs when $\mathcal{F}(h, \mathcal{D'}) \neq 0$. 
\end{definition}
(Note that $\mathcal{D'}$ can be the dataset $\mathcal{D}$ that $h$ is trained on, or another dataset $\mathcal{D}_{test} \in \mathcal{X} \times \mathcal{Y}$ in the same domain as $\mathcal{D}$).

We refer to an instance of group fairness violation as \textit{bias} and the magnitude of the bias is defined to be $|\mathcal{F}|$. If $\mathcal{F}(h, \mathcal{D}) < 0$, the learned random forest classifier $h$ is biased against the protected group. The higher the magnitude of bias, the more biased the classifier is. Typically, in machine learning applications, fairness violations of $h$ are measured on a given test dataset $\mathcal{D}_{test}$.

We are interested in identifying subsets of the training data that are \textit{responsible} for fairness violation on  $\mathcal{D}_{test}$. Training data subset $\delsubset \subseteq \mathcal{D}$ is considered responsible for an observed instance of fairness violation if removing the subset $\delsubset$ and retraining a new random forest classifier on the updated training data reduces the bias. We say subset $\delsubset$ is a \textit{responsible subset}, and formally define it below:

\begin{definition}
    \textbf{Responsible subset.} 
    Given training dataset $\mathcal{D}$ and random forest classifier $h$ trained on $\mathcal{D}$, subset $\delsubset \subseteq \mathcal{D}$ is responsible for the bias of \textit{h}'s predictions on $\mathcal{D}_{test}$, and is called a \textit{responsible subset} if 
$$|\mathcal{F}(h_{\delsubset}, \mathcal{D}_{test}| < |\mathcal{F}(h, \mathcal{D}_{test})|$$
where $h_{\delsubset}$ is the random forest classifier trained on the subset $\mathcal{D} \setminus \delsubset$ obtained after removing $\delsubset$ from $\mathcal{D}$. 
\end{definition}

To determine if a training data subset $\delsubset$ is a responsible subset, we need to quantify its contribution toward the classifier bias, which is defined as follows:

\begin{definition}
    \textbf{Subset contribution toward bias.} Given training data subset $\delsubset 
    \subseteq \mathcal{D}$ and random forest classifier $h$, the contribution of $\delsubset$ toward the bias $\mathcal{F}(h,\mathcal{D}_{test})$ of
classifier $h$ is defined as:
$$\phi_\delsubset = \dfrac{|\mathcal{F}(h_{\delsubset}, \mathcal{D}_{test})| - |\mathcal{F}(h, \mathcal{D}_{test})|}{|\mathcal{F}(h, \mathcal{D}_{test})|}$$
\label{def:contribution}
\end{definition}

The contribution of a subset toward bias essentially is the relative difference
between the bias of the original model and that of the new model
obtained by training without the subset. $-1 < \phi_\delsubset < 1$.
When $\phi_\delsubset < 0$, we say that $\delsubset$ is a responsible subset. The magnitude of $\phi_S$ indicates how much subset $\delsubset$ contributes to the bias of classifier $h$ on $\mathcal{D}_{test}$ predictions. The higher the magnitude, the higher the contribution of a subset toward bias.

Now that we have defined a responsible subset and quantified its contribution to bias, we define an \textit{explanation} as follows:
\begin{definition}
    \textbf{Explanation.} An explanation $e$ is a training data subset $\delsubset$ represented by a conjunction of literals, i.e., 
    $$e = \wedge_{j} \text{ } X_j \text{ } op \text{ } v_j$$
    where $X_j \in \mathbf{X}$, $v_j \in$ \textsf{Dom}($X_j$) and $op \in \{=, \neq, <, \le, \ge, >\}$, such that $\phi(\delsubset) < 0$.
\end{definition}

This definition of explanations identifies \textit{slices} of the training data that are responsible subsets. For Example~\ref{ex:ex1}, consider the subset $\delsubset$ = (\textsf{Age $> 45$}) $\wedge$ (\textsf{Gender=Female}). If by removing this subset from the training data and learning a new model on the updated training data, the bias of the new model is less than the bias of the orignial model, then $\delsubset$ is an explanation for the model's bias.



\noindent\textbf{Support of a subset.} We define the \textit{support} of subset $\delsubset \subseteq \mathcal{D}$ as the fraction of data instances in $\mathcal{D}$ that are contained in $\delsubset$, i.e., $sup(\delsubset) = \frac{|\delsubset|}{|\mathcal{D}|}$.

Given these preliminaries, we are interested in identifying the top-$k$ training data subsets responsible for the fairness violation of a random forest classifier on unseen test data. Formally speaking, we seek to answer the following question:

\vspace{1mm}\noindent\textbf{Problem Statement.} Given a random forest classifier $h$ trained on $\mathcal{D}$, fairness metric $\mathcal{F}$, minimum support threshold $\tau$ and parameter $k$, we address the problem of generating the top–$k$ explanations $\{{e}_i\}_{i=1}^k$ with at least $\tau$ support that have the highest contribution toward the bias in $h$'s predictions over test dataset $\mathcal{D}_{test}$. 

%


\section{Explanation Generation}
\label{sec:explanations}
In this section, we propose an algorithm to generate the top-$k$ explanations by first estimating the contribution of a subset toward bias, and then pruning the search space of subsets.

\subsection{Estimating subset contribution toward bias}
\label{sec:machine-unlearning}
The na\"ive way of computing the contribution of a subset toward model bias involves removing the subset from the training data, learning a new model with the modified training data and comparing the bias of this new model with that of the original model. However, this approach constitutes retraining the model with each subset deletion, which is a time-consuming task. 

\noindent \textbf{Machine unlearning.} To address this challenge, we we observe that learning the new model from scratch without the subset is akin to removing the effect of the subset from the trained model, and leverage recent advances in the fairly new field of machine unlearning~\cite{machine_unlearning_survey}.
The goal of machine unlearning is to \textit{unlearn} or \textit{forget} data instances in the subset by updating the trained model to completely remove the effect of the subset. The idea is to avoid retraining the model from scratch after removing data from its training dataset.


Based on the definition in~\cite{10.5555/3454287.3454603,pmlr-v139-brophy21a}, a removal method, $\mathcal{R} : \mathcal{A}(\mathcal{D}) \times \mathcal{D} \times (\mathcal{X} \times \mathcal{Y}) \rightarrow \mathcal{H}$, is a function from a model
$\mathcal{A}(\mathcal{D})$, dataset $\mathcal{D}$, and a data instance to remove from the training data $(x, y)$ to a model in $\mathcal{H}$.
For \textit{exact unlearning}, the removal method must be equivalent to applying the learning algorithm to the dataset after removing training data instance $(x, y)$. In the case of randomized learning algorithms, we define equivalence as having identical
probabilities for each model in $\mathcal{H}$ i.e., 
$P(\mathcal{A}(\mathcal{D} \setminus (x, y))) = P(\mathcal{R}(\mathcal{A}(\mathcal{D}), \mathcal{D},(x, y)))$
The na\"ive retraining approach learns a new model from scratch by retraining $\mathcal{A}$ on the modified dataset $
\mathcal{D} \setminus (x, y)$. 

In our problem setting, we are interested in removal methods that consider removal of data instances in subset $\delsubset$ i.e., 
\begin{equation}
    P(\mathcal{A}(\mathcal{D} \setminus \delsubset)) = P(\mathcal{R}(\mathcal{A}(\mathcal{D}), \mathcal{D},\delsubset))
    \label{eq:unlearning}
\end{equation}

Existing machine unlearning techniques include those specifically designed for tree-based models~\cite{schelter2021hedgecut, pmlr-v139-brophy21a}. Hedgecut~\cite{schelter2021hedgecut} focuses on improving  efficiency of the unlearning process by proposing a classification model based on extremely randomized trees (ERTs)~\cite{ERT}. The model uses a robustness quantification factor to identify \textit{robust} and \textit{non-robust} splits. The underlying assumption is that in an unlearning scenario, the structure of the tree does not change for robust splits (only some statistics are updated) while for non-robust splits, a new subtree is chosen from variants stored at the node and designated as the updated non-robust node. HedgeCut is applicable only to ERTs and assumes that a tiny fraction of data instances can be deleted (more details in~\cite{schelter2021hedgecut}). 

\noindent\textbf{Machine unlearning for random forests.} Specifically for random forests, Data Removal-Enabled Random Forests (DaRE-RF)~\cite{pmlr-v139-brophy21a} proposes a random forest variant that enables the efficient removal of training data instances. The key idea of DaRE-RF is to retrain subtrees only as needed.

To implement the unlearning process, DaRE-RF utilizes two factors--- \textit{randomness} and \textit{caching}. Randomness is introduced at the top of each tree in the forest by deciding the splitting attribute and threshold randomly. The intuition behind this decision being that since nodes near the top of the tree impact more data instances than those near the bottom, retraining them is more expensive. Since random nodes minimally depend on the statistics of the data and rarely need to be
retrained, they can improve the efficiency of unlearning.

The internal nodes cache data statistics and $k$ randomly selected thresholds for each continuous attribute. These statistics are used during unlearning. 
The leaf nodes store data instances, which ensures we know which subtrees to focus on during unlearning. When a training data instance is removed, the thresholds are checked for validity. If a threshold is invalid, the cached statistics are used to get the next valid threshold. 

In the event of unlearning, the saved statistics are sufficient to recompute the splitting criterion of each threshold without iterating through the data. Thus, by compromising the space complexity DaRE-RF greatly reduces the recomputation cost associated with unlearning. 

\noindent\textbf{Estimating subset contribution using DaRE-RF.} DaRE-RF is an \textit{exact} unlearning technique, which means that the DaRE-RF model efficiently updates the original random forest classifier after deletion of training data instances. 
Extending this idea to measuring the difference in bias on test data instances, we propose using the DaRE-RF model to compute the impact of deleting training data instances in a subset on model bias. The impact of deleting single training data instances from a trained model on the model's predictive performance using DaRE-RF has been studied~\cite{pmlr-v139-brophy21a} and shown to be within a test error difference of $1\%$. We will utilize DaRE-RF for deleting training data instances in a subset.

From
Definition~\ref{def:contribution} and Equation~\ref{eq:unlearning}, the subset contribution of subset $\delsubset$ is computed as:
\begin{equation}
    \phi_\delsubset = \dfrac{|\mathcal{F}(\mathcal{R}(\mathcal{A}(\mathcal{D}), \mathcal{D}, \delsubset), \mathcal{D}_{test})| - |\mathcal{F}(h, \mathcal{D}_{test})|}{|\mathcal{F}(h, \mathcal{D}_{test})|}
    \label{eq:dare_contribution}
\end{equation}
where $\mathcal{R}$ represents the DaRE-RF removal method.

In Section~\ref{sec:effective_unlearning_for_fairness}, we demonstrate that DaRE-RF correctly estimates the subset contribution toward bias for smaller subsets (with support $0-5\%$) and incurs a difference of $-25\%$ to $25\%$ for subsets with $5-15\%$ support.

\subsection{Pruning the subset search space}
\label{sec:pruning}
Utilizing DaRE-RF, we efficiently compute the contribution of a subset toward model bias. However, there are a large number of subsets (exponential in the number of attributes and their values), which makes this computation prohibitively expensive.  We render the problem tractable by employing several pruning techniques that reduce the subset search space.

In this section, we present the techniques \sys employs to prune the huge subset search space. Toward navigating the search space of all possible training data subsets, we borrow the concept of the apriori algorithm from frequent itemset mining~\cite{10.5555/645920.672836} and employ the lattice structure as shown in Figure~\ref{fig:lattice}. The lattice is an inverted tree-like structure where each node of the lattice corresponds to a unique subset in the training dataset represented by a conjunction of literals. Level $l$ of the lattice has subsets represented by $l$ literals. So, the top-most level (level 1) of the lattice has subsets that are represented by just one literal, level 2 has subsets with two literals and so on. The lattice is generated starting at level 1 and comprises of all feature values of the training dataset as subsets. If the training data has $n$ features with $m$ distinct values for each features, level 1 has $nm$ subsets. Nodes at level $l$ are generated by merging two nodes of level $l-1$ that have exactly $l-1$ literals in common. 

\vspace{1mm}\noindent\textbf{Greedy expansion of the lattice.}
The aforementioned process generates all possible subsets in a dataset, which are then pruned using the following rules to yield fewer subsets:

\begin{figure*}[t]
    \centering
    \includegraphics[scale=0.15,bb=-200 -100 7000 800]
    {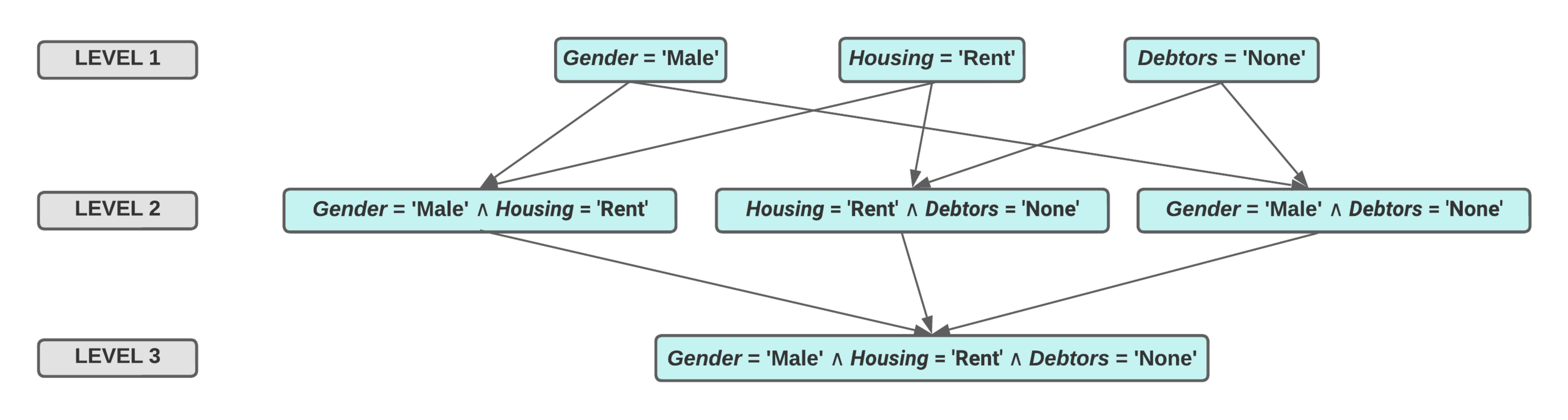}\\[-2mm]
    \caption{\textbf{Visualization of lattice structure for subset generation.} We show the first three levels of an example lattice. At level 1, all nodes consist of a single literal. For example, \textsf{Gender=`Male'} indicates data instances where the column \textsf{Gender} has the value \textsf{`Male'}. At each level, literals are merged two at a time, as illustrated, to generate subsequent subsets.
    }
    \label{fig:lattice}
    \vspace{-2mm}
\end{figure*}

\noindent \textbf{Rule 1: Prune irrelevant subsets.} While generating subsets by navigating the lattice structure, we ensure that impractical subsets such as \textsf{(Age $< 50$) $\wedge$ (Age $> 70$)} are not generated. Upon expanding the levels in the lattice, progressively complex subsets (having more literals) are generated that have smaller support compared to subsets higher up in the lattice.

\noindent \textbf{Rule 2: Filter subsets depending upon a support threshold.} Depending on the domain, we might want the responsible subsets to lie within some specific \textit{support} range only. For example, for one domain it might be beneficial to identify larger responsible subsets while for another, we might want to inspect smaller responsible subsets. Both of these scenarios are interesting in their own right and highlight different issues (potential systemic bias in the former and possible errors in the latter). This choice is especially important while deleting data instances corresponding to such subsets. Deleting subsets having large {support} may be undesirable due to the large reduction in training dataset size it causes. While expanding the lattice structure, if we encounter a node such that the subset represented by it has a {support} smaller than the minimum {support} threshold, then we do not expand its subtree as its children (with more literals and stricter conditions) will have an even smaller support value. In case the support of a subset is more than the maximum user-defined support level, we consider it for further expansion of the lattice structure as it may generate subsets in the required support range in subsequent levels, but exclude it from our explanations.

\noindent\textbf{Rule 3: Prune complex subsets.} The complexity of a subset is indicated by its interpretability which is directly indicated by the number of literals used to represent it. A subset represented by three to four literals is way more comprehensible than one having ten or more literals. Limiting the number of literals desired in the responsible subsets helps us implement a stopping condition for expanding the lattice structure. We keep expanding the levels of the lattice structure until we reach a level that has subsets represented by the maximum desired number of literals as defined by a user of the system.

\noindent\textbf{Rule 4: Prune subsets with lower contribution to bias than its parent subsets.} Even with the aforementioned pruning techniques, the subset search space could still be large. We need to ensure that
no redundant branches of the lattice structure are traversed that might result in expanding subsets that are not responsible.
Consider subset $S$ representing a node in the lattice structure. Let $S_1$ and $S_2$ represent the subsets that $S$ was merged from in the lattice structure. Let $\phi_S$, $\phi_{S1}$ and $\phi_{S2}$ represent the subset contribution of $S$, $S_1$ and $S_2$ respectively computed using Equation~\ref{eq:dare_contribution}. If the subset contribution of $S$ is lower than either of its parents $S_1$ and $S_2$, i.e., $\phi_S < \phi_{S1}$ or $\phi_S < \phi_{S2}$, then $S$ is considered to be of a worse quality and the node representing subset $S$ is not expanded further. The intuition behind this rule is that a subset with a lower contribution to bias is devoid of more powerful parts of the training data (present in the subsets represented by its parents), and hence is not considered a beneficial route to follow. (In our experiments, we considered two strategies for comparing the contribution of a subset with that of its parents. The \textit{normal} strategy compares the subset contribution of the child and parent nodes directly, while the \textit{per instance} strategy compares the subset contributions over the corresponding subset sizes. The latter is a way of normalizing the contributions for subsets of differing sizes).

\noindent\textbf{Rule 5: Prune subsets that are not responsible.} This rule ensures that a node in the lattice structure is expanded only if the subset contribution of the subset $S$ represented by the node is positive, i.e., $\phi_S > 0$. The intuition behind this rule is that we are only interested in subsets that, when removed from the training data, improve upon the original model's fairness. 


\vspace{1mm}To implement the aforementioned pruning rules, we introduce the following hyperparameters: 
\begin{enumerate}
    \item {\textsf{compareStrategy:}} \textit{perInstance} if  bias reduction of a subset should be averaged over subset size; \textit{normal}, otherwise.
    \textit{perInstance} enables comparing subsets with substantially different sizes.
    \item \textsf{{compareOriginalParity:}} (boolean) \textsf{true} requires that bias reduction of a newly merged subset in lattice structure be greater than zero, ensuring that removing the subset results in learning a fairer model.
    \item {\textsf{maxLiterals: }} indicates the maximum number of literals the user wants in the top-$k$ explanations.
    \item {\textsf{supportRange: }} indicates the desired range for support of the reposnible subsets.
\end{enumerate}

\subsection{Putting it all together}
With the subset contribution computation phase using DaRE-RF and the phase that prunes the subset search space using the lattice structure, \sys intertwines these two phases to generate top-$k$ explanations. First, it expands the lattice structure in a top-down manner, generates subsets at levels 1 and 2, and computes their subset contribution. These subsets are \textit{candidates} for explanations, and are then pruned using the pruning rules in Section~\ref{sec:pruning}. Subsets represented by nodes in the subsequent levels are then generated in accordance with the pruning rules. The algorithm stops when either the maximum depth of the lattice is reached (controlled by Rule 3) or there are no more subsets within the desired support range. The candidate subsets are then sorted in decreasing order of their contribution to bias, and the top-$k$ subsets are with the highest contribution toward bias are generated as explanations. Algorithm~\ref{alg:algo1} presents the pseudocode for the explanation generation process.




    

        



\begin{algorithm}
    \caption{Explanation Generation}
    \label{alg:algo1}
    \KwInput{maxLiterals $\eta$, supportRange $\tau$, compareStrategy $\alpha$, compareOriginalParity $\beta$ }
    \KwOutput{Top-$k$ explanations $\mathcal{E} = \{e_i\}_{i=1}^k$}
        
        $N$ = [ ]
        
        
        $Level = 1$
        

        $E \gets ExpandSubsets(N, Level)$ \Comment{Rule 1}
        
        
        $\mathcal{E}$ = [ ]
        
        \Comment{Rule 3}

        \While{$ Level <= \eta $}{
            \For{subset in E}{ \Comment{Rule 2}{
                
                \If{\textbf{NOT} $IsValidSupport(subset, \tau)$}{
                    \If{$sup(subset) > MAX(\tau)$} {
                        $N.append(subset)$
                    }
                    \Else{CONTINUE}
                }\Comment{Rules 4 and 5}
                
                $selected \gets EvaluateSubset(subset, \alpha, \beta)$
                
                \If{$selected$ == \textbf{TRUE}} {
                    $N.append(subset)$
                    $\mathcal{E}.append(subset)$
                }
            }
            $L = L + 1$ \Comment{Rule 1}
            
            $E \gets ExpandSubsets(N, Level)$

            \If{\textbf{NOT} $E$}{ 
                \textbf{BREAK} 
            }    
        } 
        \Return $\mathcal{E}$
    }
\end{algorithm}
\section{Experimental Evaluation}
\label{sec:experiments}

In this section, we answer the following research questions.
\textbf{RQ1:} How effective are machine unlearning approaches in capturing the effect of subset removal on fairness? \textbf{RQ2:} How effective and interpretable are the explanations generated by \sys? \textbf{RQ3:} How efficient is \sys's generation of data-based explanations?

\subsection{Experimental Setup}
\subsubsection{Datasets}
We demonstrate the effectiveness of \sys on the following standard datasets in the fairness in machine learning literature:

\noindent \textbf{German Credit~\cite{Dua2019}.}  This dataset contains financial information of $1000$ individuals, the sensitive attribute is “age”, and the prediction task is to determine whether an individual is a good credit risk or a bad credit risk.

\noindent \textbf{Census Income~\cite{adult}.} This dataset contains demographic and financial information of $48,844$ individuals, the  sensitive attribute is “sex” 
and the prediction task determines whether an individual has annual income $\leq50k$ or $>50k$. 

\noindent \textbf{SQF~\cite{sqf-data}.} The Stop-Question-Frisk dataset contains demographic and stop-related information for
$72,548$ individuals who were stopped and questioned (and possibly frisked) by the NYC Police Department (NYPD). The classification task is
to predict if a stopped individual will be frisked.

\noindent More details on these datasets can be found in Table~\ref{table:datasets}. As illustrated by the distribution of dataset population between the protected and the privileged groups over their count and fraction of positive labels for all the datasets, it is clearly seen that the proportion of positive labels for the privileged group is more than that of the protected group. This imbalance gives an indication of a potential disparity amongst the two groups.

\begin{table*}[t]
\captionsetup{justification = centering, labelsep = newline}
\caption{\textbf{Summary of datasets}. 
Pri\_count\% = \% of privileged group, Pro\_count\% = \% of protected group, Pri\_pos\% = privileged group positive label \%, Pro\_pos\% = protected group positive label\%}
 \label{table:datasets}
 \centering
 \begin{tabular}{|c|c|c|p{1.4cm}|c|l|l|l|} \hline 
      \textbf{Dataset}& \textbf{\# instances}&  \textbf{\# features}& \textbf{Sensitive attribute}& \textbf{Pri\_count\%}& \textbf{Pro\_count\%}& \textbf{Pri\_pos\%}& \textbf{Pro\_pos\%}\\ \hline 
      German Credit &  1000&  21& age&58.90\%& 41.10\%& 74.19\%&63.99\%
\\ \hline 
 Adult Income & 45,222& 10& sex& 67.50\%& 32.50\%& 31.24\%&11.35\%
\\\hline 
      Stop Question Frisk&  72,546&  16& race&64.06\%& 35.94\%& 38.32\%&30.16\%
\\\hline
 \end{tabular}
\end{table*}

\subsubsection{Metrics} We report the parity reduction to reflect the improvement in the fairness of the original and the updated model, while accuracy reduction reports the decrease in the accuracy of the updated model compared to the original model.

\vspace{1mm}\noindent\textbf{Fairness Metrics: } As discussed in Section~\ref{sec:notations}, \sys supports three fairness metrics: statistical parity difference, predictive parity difference and equalizing odds parity difference~\cite{10.1145/3457607,verma2018fairness}.
We find the fairness metric difference between the privileged and protected groups and use that as our fairness criteria. A difference of 0 indicates an unbiased model. If the difference is not zero, then the model is biased against one of the groups.

\vspace{1mm}\noindent\textbf{Bias Reduction: } For a training data subset, it is the percent by which the fairness metric difference of the model reduced after removing the subset from the training dataset.


\noindent\textbf{Hardware and Platform}. The experiments were conducted on a 64-bit Windows OS with a 3.20 GHz AMD Ryzen 7 5800H processor and 32.0 GB memory. The algorithms were implemented in Python in Jupyter Notebook environment. 

\vspace{1mm}\noindent\textbf{Source code.} The source code for our system is publicly available at the following  repository:~\url{https://github.com/tanmaysurve/FairDebugger}


\subsection{Effectiveness of machine unlearning for debugging fairness} \label{sec:effective_unlearning_for_fairness}

The effect of DaRE-RF unlearning capabilities on accuracy is known~\cite{pmlr-v139-brophy21a}: as the \textit{support} of a subset to be unlearned from the model increases, the accuracy of the updated model decreases. 
To test the applicability of DaRE-RF's unlearning to the use-case of fairness debugging, we first need to validate its impact on model fairness.
\begin{figure*}[t]
    \centering
    \includegraphics[scale=0.5,bb=150 200 900 650]
    {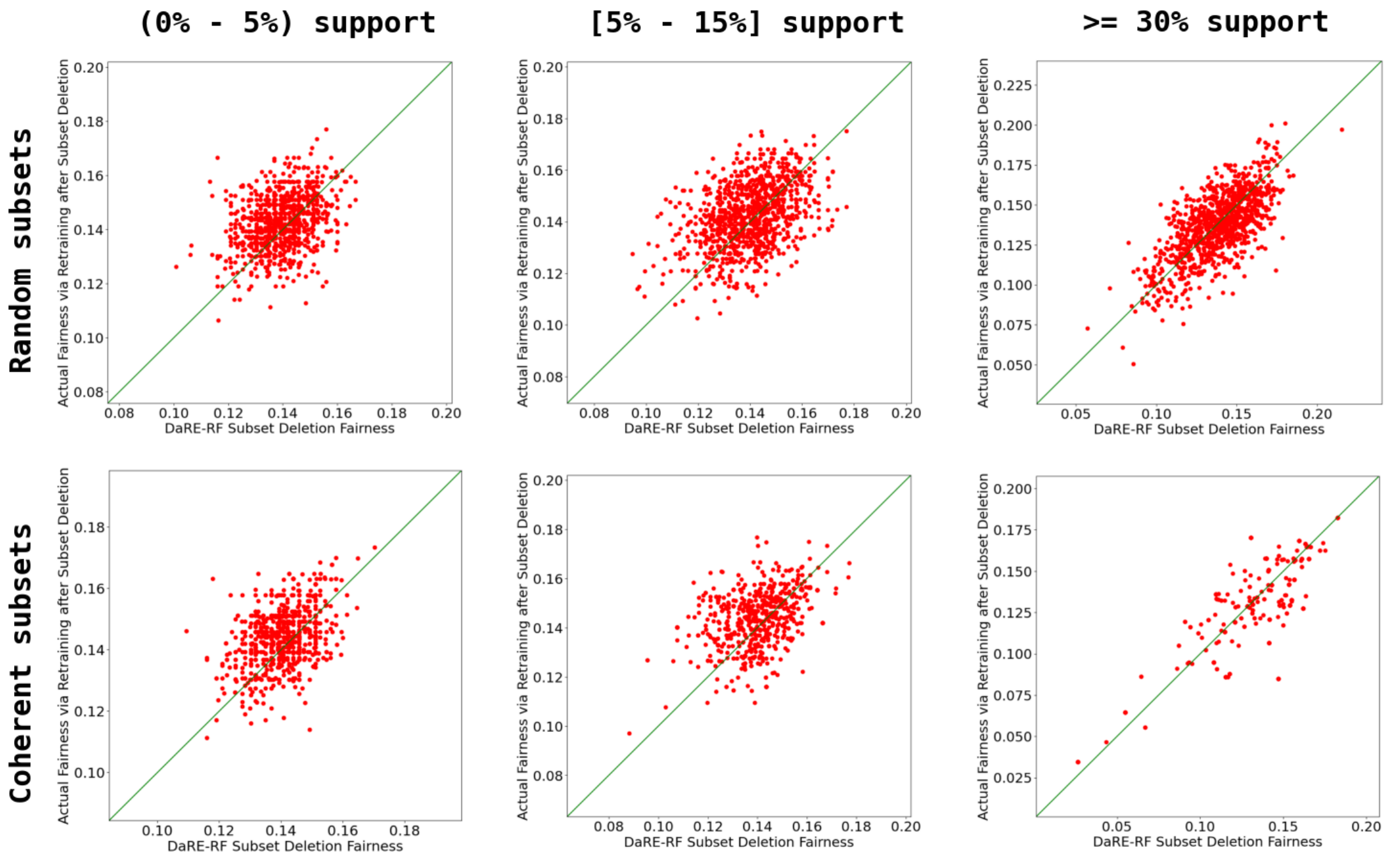}
    \\
    \vspace{40mm}
    \caption{\textbf{Effect of DaRE-RF's unlearning capability} on fairness for 1000 \textit{random} subsets and 1000 \textit{coherent} subsets. The top 3 plots correspond to \textit{random} subsets for 3 subset \textit{support} ranges while the bottom 3 plots correspond to \textit{coherent} subsets. Plots correspond to the predictive parity fairness metric. x-axis indicates model fairness after DaRE-RF's unlearning was used to remove influence of a subset from the model. y-axis corresponds to fairness of model retrained after removing the said subset from the training data. The green line is $y = x$ line. }
    \label{fig:dare_rf_effect_on_fairness}
\end{figure*}
Toward this goal, we use 1,000 \textit{random} and 1,000 \textit{coherent} subsets from the German credit dataset \cite{Dua2019}. A \textit{random} subset is a subset in which the data instances are chosen randomly while a \textit{coherent} subset refers to one defined in the form of conjunction of literals. We consider three {support} ranges: ($0\% - 5\%$), $[5\% - 15\%]$ and $\ge 30\%$, and three fairness metrics: statistical parity, predictive parity and equalizing odds.
Figure \ref{fig:dare_rf_effect_on_fairness} shows the plots for \textit{random} and \textit{coherent} subsets for predictive parity fairness metric across all three \textit{support} ranges. We see that most of the points in these plots lie on or near the green color $y = x$ line. This means that the $\mathcal{F}$ calculated via retraining from scratch after deleting a subset from training dataset and calculated after using unlearning to remove influence of the same subset is almost the same. Unlike accuracy, 
fairness of the unlearned model is approximately the same as that of the retrained model even with increasing subset size. 
This experiment highlights that model fairness is preserved by DaRE-RF unlearning across various \textit{support} ranges. Similar results were obtained for statistical parity and equalizing odds fairness metrics. Consequently, we can use DaRE-RF for our algorithm as it would not adversely affect fairness. 

\begin{table*}
\captionsetup{justification = centering, labelsep = newline}
\centering
\caption{Top-5 explanations for German Credit dataset in the support range 5\% - 15\% for statistical parity fairness metric.}
 \begin{tabular}{|c|p{8cm}|c|c|c|} \hline 
      \textbf{Index}&  \textbf{Patterns}&  \textbf{Support}& \textbf{Parity Reduction} &\textbf{Accuracy Reduction}\\ \hline 
      GS1&  status\_chec\_acc = `$<0$ DM', num\_people\_liable\_to\_maint = 'high'&  5.00\%& 97.79\%&3.42\%\\ \hline 
      GS2&  savings = '100 $\leq$ .. $<$ 500 DM', job = 'skilled employee / official'&  7.13\%& 95.58\%&1.37\%\\ \hline 
      GS3&  install\_plans = 'bank', debtors = 'none'&  12.00\%& 93.38\%&0.68\%\\ \hline 
      GS4&  status\_chec\_acc = `no checking account', property = `unknown / no property'&  5.25\%& 91.17\%&0\%\\ \hline 
      GS5&  housing = `rent', status\_and\_sex = `female: divorced/separated/married'&  10.00\%& 89.91\%&2.74\%\\ \hline
 \end{tabular}
 \label{table:top5PatternsGermanCreditSP}
\end{table*}

\begin{table*}
\captionsetup{justification = centering, labelsep = newline}
 \centering
  \caption{Top-5 explanations for German Credit dataset in the support range 5\% - 15\% for predictive parity fairness metric.}
 \begin{tabular}{|c|p{8cm}|c|c|c|} \hline 
      \textbf{Index}&  \textbf{Patterns}&  \textbf{Support}& \textbf{Parity Reduction} &\textbf{Accuracy Reduction}\\ \hline 
      GP1&  age = 'old', job = 'management / self-employed / highly qualified employee / officer'&  10.00\%& 25.28\%&-1.37\%\\ \hline 
      GP2&  age = 'old', cred\_hist = 'delay in paying off in the past'&  5.38\%& 23.68\%&-2.05\%\\ \hline 
      GP3&  employment = '1 $<=$ .. $<$ 4 years', cred\_hist = 'critical account / other credits existing (not at this bank)'&  8.13\%& 20.78\%&-0.68\%\\ \hline 
      GP4&  job = 'management / self-employed / highly qualified employee / officer', intallment\_rate = 'high'&  7.50\%& 19.68\%&-1.37\%\\ \hline 
      GP5&  job = 'management / self-employed / highly qualified employee / officer', foreign\_worker = 'yes'&  14.00\%& 19.27\%&-0.68\%\\ \hline
 \end{tabular}
 \label{table:top5PatternsGermanCreditPP}
\end{table*}

\begin{table*}
\captionsetup{justification = centering, labelsep = newline}
 \centering
 \caption{Top-5 explanations for German Credit dataset in the support range 5\% - 15\% for equalizing odds parity fairness metric.}
 \vspace{-1mm}
 \begin{tabular}{|c|p{8cm}|c|c|c|} \hline 
      \textbf{Index}&  \textbf{Patterns}&  \textbf{Support}& \textbf{Parity Reduction} &\textbf{Accuracy Reduction}\\ \hline 
      GE1&  cred\_amt = 'high', status\_chec\_acc = '$<$ 0 DM'&  6.38\%& 100\%&2.74\%\\ \hline 
      GE2&  status\_and\_sex = 'female: divorced/ separated/ married', present\_resi\_since = 'high'&  14.25\%& 100\%&2.74\%\\ \hline 
      GE3&  present\_resi\_since = 'high', employment = '1 $<=$ .. $<$ 4 years'&  10.63\%& 100\%&-1.37\%\\ \hline 
      GE4&  duration = 'high', job = 'unskilled - resident'&  6.25\%& 100\%&2.73\%\\ \hline 
      GE5&  property = 'unknown / no property', duration = 'low'&  6.00\%& 100\%&1.23\%\\ \hline
 \end{tabular}
 \label{table:top5PatternsGermanCreditEO}
 \vspace{-2mm}
\end{table*}

\subsection{Effectiveness of \sys's explanations}


We analyze the top-$k$ explanations by evaluating how effective they are in explaining instances of group fairness violations. We are interested in understanding if the underlying training data can justify the explanations.
Toward this goal, we use the following two methods to analyze the explanations:
\begin{enumerate}
    \item {Sensitive groups label proportions: } We check the proportion of positive outcomes in the training data for both the sensitive groups in an explanation. A very high positive prediction for the privileged group compared to the protected group in a subset is a plausible reason for the presence of the subset in the top-$k$ explanations.
    
    \item {Model feature importance deviations: } When a subset is deleted from a learned model, there is a possibility that the feature importance rankings change. Comparing the feature importance ranking of a model with and without a particular training data subset points to its  impact on model predictions. If the importance of the sensitive attributes (e.g., \textit{gender}) reduces after deleting the subset from the training data, it explains why the subset was featured in the top-$k$ explanations generated by \sys.
\end{enumerate}

Sensitive groups label proportions along with model feature importance deviations provide good enough evidences as to why any pattern generated is actually a bias-inducing pattern. These tests or methods in a way also act as a validation tool for our algorithm as they provide empirical evidence towards the fairness-based bias-inducing effect of the resultant patterns.

\begin{table*}
\captionsetup{justification = centering, labelsep = newline}
 \centering
 \caption{Top-5 explanations for Adult Income dataset in the support range 5\% - 15\% for statistical parity fairness metric.}
 \vspace{-1mm}
 \begin{tabular}{|c|c|c|c|c|} \hline 
      \textbf{Index}&  \textbf{Patterns}&  \textbf{Support}& \textbf{Parity Reduction} &\textbf{Accuracy Reduction}\\ \hline 
      AS1&  sex = 'Male', education = 'Bachelors'&  11.67\%& 51.89\% &-1.47\%\\ \hline 
      AS2&  occupation = 'Sales', age = 'Middle-aged'&  6.54\%& 36.43\% &-1.04\%\\ \hline 
      AS3&  occupation = 'Adm-clerical'&  12.33\%& 35.53\% &1.90\%\\ \hline 
      AS4&  age = 'Middle-aged', workclass = 'Self-emp-not-inc'&  6.01\%& 34.39\% &2.34\%\\ \hline 
      AS5&  relationship = 'Unmarried'&  10.64\%& 34.37\% &1.13\%\\ \hline
 \end{tabular}
 \label{table:top5PatternsAdultIncomeSP}
\end{table*}

\begin{table*}
\captionsetup{justification = centering, labelsep = newline}
 \centering
 \caption{Top-5 explanations for Adult Income dataset in the support range 5\% - 15\% for predictive parity fairness metric.}
 \begin{tabular}{|c|c|c|c|c|} \hline 
      \textbf{Index}&  \textbf{Patterns}&  \textbf{Support}& \textbf{Parity Reduction} &\textbf{Accuracy Reduction}\\ \hline 
      AP1&  sex = 'Female', education = 'Bachelors'&  5.04\%& 2.32\% &-0.15\%\\ \hline 
      AP2&  occupation = 'Craft-repair', education = 'HS-grad'&  6.30\%& 2.26\% &0.42\%\\ \hline 
      AP3&  occupation = 'Craft-repair'&  13.36\%& 1.97\% &0.49\%\\ \hline 
      AP4&  hours-per-week = 'Full-time', education = 'Bachelors'&  9.03\%& 1.70\% &-0.52\%\\ \hline 
      AP5&  education = 'Some-college', relationship = 'Husband'&  7.85\%& 1.56\% &0.41\%\\ \hline
 \end{tabular}
 \label{table:top5PatternsAdultIncomePP}
\end{table*}

\begin{table*}
\captionsetup{justification = centering, labelsep = newline}
 \centering
 \caption{Top-5 explanations for Adult Income dataset in the support range 5\% - 15\% for equalizing odds parity fairness metric.}
 \begin{tabular}{|c|p{8cm}|c|c|c|} \hline 
      \textbf{Index}&  \textbf{Patterns}&  \textbf{Support}& \textbf{Parity Reduction} &\textbf{Accuracy Reduction}\\ \hline 
      AE1&  marital-status = 'Married-civ-spouse', occupation = 'Exec-managerial'&  7.94\%& 95.86\%&-1.95\%\\ \hline 
      AE2&  age = 'Middle-aged', occupation = 'Exec-managerial'&  9.96\%& 95.54\%&-2.03\%\\ \hline 
      AE3&  occupation = 'Exec-managerial', race = 'White'&  11.91\%& 95.08\%&-1.94\%\\ \hline 
      AE4&  age = 'Middle-aged', education = 'Some-college'&  12.31\%& 90.19\%&-1.26\%\\ \hline 
      AE5&  occupation = 'Exec-managerial'&  13.23\%& 88.01\%&-1.96\%\\ \hline
 \end{tabular}
 \label{table:top5PatternsAdultIncomeEO}
\end{table*}

\begin{table*}
\captionsetup{justification = centering, labelsep = newline}
 \centering
 \caption{Top-5 explanations for Stop-Question-Frisk dataset in the support range 5\% - 15\% for statistical parity. cs\_casng = reason for stop – casing a victim (0) or location (1), cs\_descr = reason for stop – fits a relevant description (1), cs\_lkout = reason for stop – suspect acting as a lookout (1), cs\_drgtr = reason for stop – actions indicative of a drug transition (1).}
 \begin{tabular}{|c|c|c|c|c|} \hline 
      \textbf{Index}&  \textbf{Patterns}&  \textbf{Support}& \textbf{Parity Reduction} &\textbf{Accuracy Reduction}\\ \hline 
      SS1&  sex\_M = '0'&  6.51\%& 100\%&1.78\%\\ \hline 
      SS2&  weight = 'light', cs\_casng = '0'&  6.44\%& 39.95\%&0.62\%\\ \hline 
      SS3&  build = 'heavy', cs\_descr = '0'&  6.87\%& 35.99\%&0.04\%\\ \hline 
      SS4&  cs\_lkout = '0', cs\_drgtr = '1'&  6.01\%& 33.83\%&-0.17\%\\ \hline 
      SS5&  weight = 'light'&  7.81\%& 31.24\%&1.21\%\\ \hline
 \end{tabular}
 \label{table:top5PatternsFriskSP}
\end{table*}

\begin{table*}
\captionsetup{justification = centering, labelsep = newline}
 \centering
 \caption{Top-5 explanations for Stop-Question-Frisk dataset in the support range 5\% - 15\% for predictive parity fairness metric. cs\_objcs = for stop – carrying suspicious object (1)}
 \label{table:top5PatternsFriskPP}
 \begin{tabular}{|c|c|c|c|c|} \hline 
      \textbf{Index}&  \textbf{Patterns}&  \textbf{Support}& \textbf{Parity Reduction} &\textbf{Accuracy Reduction}\\ \hline 
      SP1&  sex\_M = '0'&  6.51\%& 100\%&1.78\%\\ \hline 
      SP2&  weight = 'light', cs\_objcs = '0'&  7.57\%& 87.24\%&1.20\%\\ \hline 
      SP3&  cs\_objcs = '0', age = '$>$ 60 yrs'&  10.74\%& 53.06\%&0.53\%\\ \hline 
      SP4&  weight = 'light'&  7.81\%& 42.12\%&1.21\%\\ \hline 
      SP5&  age = '$>$ 60 yrs'&  11.49\%& 42.10\%&0.36\%\\ \hline
 \end{tabular}
\end{table*}

\begin{table*}
\captionsetup{justification = centering, labelsep = newline}
 \centering
 \caption{Top-5 explanations for Stop-Question-Frisk dataset in the support range 5 - 15\% for equalizing odds parity fairness metric. cs\_casng = reason for stop – casing a victim (0) or location (1), cs\_descr = reason for stop – fits a relevant description (1), cs\_lkout = reason for stop – suspect acting as a lookout (1), cs\_drgtr = reason for stop – actions indicative of a drug transition (1), perobs = period of observation, inout\_O = stop was outside (1).}
 \begin{tabular}{|c|c|c|c|c|} \hline 
      \textbf{Index}&  \textbf{Patterns}&  \textbf{Support}& \textbf{Parity Reduction} &\textbf{Accuracy Reduction}\\ \hline 
      SE1&  sex\_M = '0'&  6.51\%
& 100\%&1.78\%\\ \hline 
      SE2&  build = 'heavy', cs\_descr = '0'&  6.86\%& 43.98\%&0.03\%\\ \hline 
      SE3&  cs\_lkout = '0', cs\_drgtr = '1'&  6.01\%& 42.69\%&-0.17\%\\ \hline 
      SE4&  weight = 'light', cs\_casng = '0'&  6.44\%& 42.48\%&0.62\%\\ \hline 
      SE5&  perobs = '0.0', inout\_O = '0'&  12.34\%& 27.98\%&0.04\%\\ \hline
 \end{tabular}
 \label{table:top5PatternsFriskEO}
\end{table*}

\vspace{1mm}\noindent \textbf{Explanations for \textsf{Adult: }} From Tables \ref{table:top5PatternsGermanCreditSP}, \ref{table:top5PatternsGermanCreditPP} and \ref{table:top5PatternsGermanCreditEO}, we can clearly see that subsets GS1 - GS5 and GE1 - GE5 corresponding to statistical parity and equalizing odds respectively outperforms subsets GP1 - GP5 corresponding to predictive parity in terms of parity reduction. GS1 - GS5 removes almost all bias and GE1 - GE5 removes 100\% of the model bias corresponding to their respective fairness metrics. GP1-GP5 could only approach a max 25.28\% parity reduction. Though not very bad, compared to others it falls way behind.
On the other hand, we can easily see that accuracy reduction is not very high, lying below 3\% in all cases. In some cases, like, GP1 – GP5 and GE3 ended up improving overall accuracy of model when their influence was removed from the model.

The subsets corresponding to different fairness metrics do not show much resemblance to each other. This highlights that there is no particular subset or data points that induces high bias corresponding to multiple fairness metric at the same time.

We find that GS1 and GS2 can be explained by having larger positive label percent for their privileged group, old people, (70\% and 70.1\% compared to 57\% and 62\% for protected group respectively). On the other hand, for subsets GP3 and GP4 the positive label proportion for the protected group is higher. Hence these cannot be explained by just looking at the positive label proportions of sensitive groups. Considering feature importance deviation, we found that removing GP3 and GP4 improves relevant features like status of checking account by 12.08\% and 9.87\% respectively. The savings attribute’s importance also increased by 7.27\% and 17.49\% for GP3 and GP4 respectively. On the other hand, the importance of discriminatory feature like been a female: divorced/separated/married decreased by a whopping 38.55\% and 34.20\% for B3 and B4 respectively. GE1 - GE5 all have higher positive label percentage for their privileged group and can thus be explained this way.
Thus, all the subsets in this way are interpretable and are effective, albeit, some more than others.

\vspace{1mm}\noindent \textbf{Explanations for \textsf{German: }} From Tables \ref{table:top5PatternsAdultIncomeSP}, \ref{table:top5PatternsAdultIncomePP} and \ref{table:top5PatternsAdultIncomeEO}, we find similar trends for parity reduction as that for the German credit dataset. The parity reduction for subsets AP1 – AP5 corresponding to predictive parity is extremely low (highest having 2.32\%) and not effective enough to remove the bias. AE1 – AE5 subsets are extremely bias inducing having high parity reduction (88\% - 96\%). The subsets corresponding to statistical parity (AS1 – AS5) have good enough parity reduction (34\% - 52\%) to be called effective.

The accuracy reduction again is very small ($< 3\%$) across all subsets. AE1 – AE5 even increase the accuracy of model by $1\% - 2\%$. Unlike in the case of German credit dataset in previous section, the subsets for Adult Income dataset across different fairness metric have a lot of common features. (e.g., AS1 and AP1 both have education = ‘Bachelors’ pattern). Many attributes are common across all the subsets even though their values may be different. This indicates the importance of such attributes in inducing bias in the model.

Considering the interpretability/explainability of the explanations, we find that none of the subsets are explainable by looking at positive label percent for sensitive groups as they are close enough to generate any clear disparity. Upon deletion of AS2 and AS3, the feature importance of sensitive attribute sex dropped by 34.42\% and 34.87\% respectively and the importance of relevant attribute for income like occupation increased by 30.53\% and 41.79\% respectively. Subsets AP4 and AP5 tell similar story as we find that their deletion reduced feature importance of both sensitive attributes sex and age by (15.60\% and 24.66\%) and (17.89\% and 23.27\%) respectively. On the other hand, feature importance of highly important features like education and occupation went up by the range of [5\% - 10\%] in both. Similarly, for AE3 and AE4 the importance of discriminatory and less relevant features like age and sex decreased while that of education and workclass (by around 49\% and 24\% respectively) increased. Baring the non-effective subsets for predictive parity, \sys produced very effective and interpretable subsets.

\vspace{1mm}\noindent \textbf{Explanations for \textsf{SQF: }}All the subsets across fairness metrics provide good parity reduction in Tables \ref{table:top5PatternsFriskSP}, \ref{table:top5PatternsFriskPP} and \ref{table:top5PatternsFriskEO}. Unlike the Adults income and German credit dataset, we have effective bias reducing subsets SP1 – SP5 corresponding to predictive parity. A very interesting thing to note here is that each fairness metric bias is reduced by 100\% upon removal of the subset sex\_M = ‘0’, which corresponds to removing all females from the dataset. Another thing of note here is the similarities between subsets of different fairness metric. All the subsets across fairness metrics consist of similar attributes and their corresponding values.
The accuracy reduction is again very small ($< 2\%$) across all subsets and deletion of some subsets (SS4, SE3) even improved the accuracy a little.

In terms of subset \textit{interpretability}, again as in the Adults dataset, positive label percent comparison between sensitive groups does not provide much of an explanation as they are quite similar across sensitive groups for all the subsets.
Studying the feature importance changes after deletion of these subsets, provides a clear explanation. Deleting SS1 and SS5 caused the importance of feature cs-drgtr (check table captions for meaning of attribute abbreviations) to increase by 116.23\% and 58.46\% respectively. Similarly, cs-objcs importance increased by 75.30\% and 42.50\% for SS1 and SS5, respectively. Importance of cs-lkout also increased to 118.74\% and 32.45\% for SS1 and SS5, respectively. These features are valid and relevant reasons to perform the SQF procedure. The highest loss in feature importance was seen in sex-M, dropping by 100\% and 30.45\% in SS1 and SS5, respectively (this is good since gender should not be an important criteria in determining whom to frisk). The other subsets also follow similar trends.
\sys generated effective subsets for all fairness metrics for this dataset unlike the other datasets where subsets corresponding to predictive parity were less effective. Moreover, all the subsets can be easily interpreted by finding the feature importance deviations before and after deletion of a subset from the model.

\subsection{Efficiency of \sys}
We observed from the previous sections that \sys provides quite effective and interpretable data-based explanations. However, it is also important to consider how efficient it is in generating the explanations. We measure efficiency by reporting the runtime of \sys, and consider how it is impacted by the dataset dimensions ($n \times m$ where $n$ represents the number of data instances and $m$ is the number of features in the dataset.
\begin{table}[h]
\captionsetup{justification = centering, labelsep = newline}
 \centering
 \caption{Runtime of \sys for different dataset for 3 fairness metrics. SP-time, PP-time and EO-time are \sys runtime in seconds for statistical parity, predictive parity, and equalizing odds parity respectively. Dimension refers to number of data instances x number of features of the dataset. }
 \begin{tabular}{|c|c|c|c|c|} \hline 
      \textbf{Dataset}&  \textbf{Dimension}&  \textbf{SP-time (sec)}& \textbf{PP-time (sec)}&\textbf{EO-time (sec)}\\ \hline 
      \textit{German credit}&  21,000 \textbf{(1x)}&  130.96 \textbf{(1x)}& 132.74 \textbf{(1x)}&141.20 \textbf{(1x)}\\ \hline 
      \textit{Adult income}&  452,220 \textbf{(21.5x)}&  687.92 \textbf{(5.3x)}& 616.24 \textbf{(4.6x)}&629.41 \textbf{(4.4x)}\\ \hline 
      \textit{SQF}&  1,160,736 \textbf{(55x)}&  5268.25 \textbf{(40.2x)}& 5329.75 \textbf{(40x)}&5144.25 \textbf{(36.4x)}\\\hline
 \end{tabular}
 \label{table:timeTakenRealDataset}
\end{table}

\begin{figure}[h]
    \centering
    \includegraphics[width=.6\columnwidth]
    {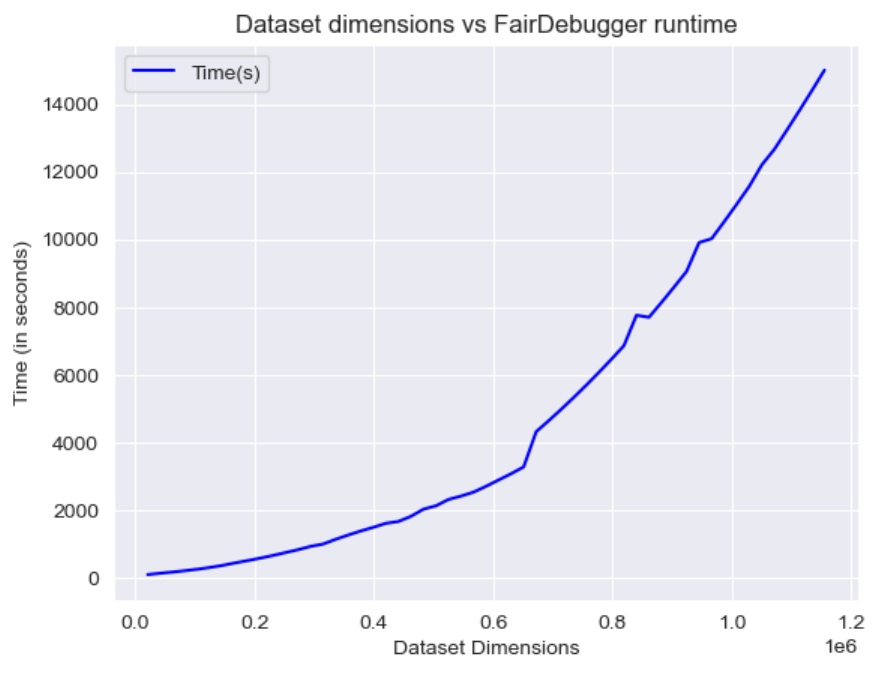}\\[-2mm]
    \caption{Runtime of \sys for various dataset dimensions. Dimension refers to num\_of\_data\_instances $\times$ num\_of\_attributes. We observe that the runtime increases almost quadratically as dataset dimension increases.}
    \label{fig:dare_rf_time_syn_data}
\end{figure}

Table \ref{table:timeTakenRealDataset} shows that as dataset dimension increases exponentially from German credit to Adult income by a factor of 21.5x, the runtime of \sys increases quite linearly with an average 4.7x across all fairness metrics. The factor of increase of \sys’s runtime is 4.5 times lesser than factor of increase in dataset dimension.  The dataset dimension again increases exponentially from Adult income to SQF dataset by a factor of 2.5x while the runtime of \sys increases even more by a factor of 8.3x on an average over all fairness metrics. 
This indicates that \sys works efficiently for small to medium dimensions dataset but starts having exponential increase in runtime as dataset dimensions increases further. To verify this further, \sys was used on synthetic datasets of increasing dimension to check the trend of its runtime with respect to dataset dimensions. Fig. \ref{fig:dare_rf_time_syn_data} shows that \sys is very efficient for datasets of small to medium dimensions as indicated by smaller slope in the beginning of the graph. After around 630,000 dataset dimension, the slope starts increasing and \sys's runtime keeps increasing exponentially. In general, the runtime increases quadratically with increasing dataset dimensions.

\section{Related Work}
Our work is broadly related to the following lines of recent research:
\noindent\textbf{Explainable AI.} Our research is mainly related to the broad field of explainable artificial intelligence~\cite{arrieta2020explainable,guidotti2018survey,molnar2020interpretable} that is aimed at ensuring that the decisions made by an AI-based system are transparent to and understandable by the different stakeholders of the system (e.g., end-users impacted by its decisions, businesses deploying the system, and practitioners using off-the-shelf ML packages for making those decisions). XAI techniques primarily generate explanations for model decisions in terms of {features} or {examples} of the underlying training data. The key idea in \textit{feature-based} explanations~\cite{DBLP:conf/nips/LundbergL17,ribeiro2016should,ribeiro2018anchors,galhotra2021explaining,mothilal2020explaining,wachter2017counterfactual} is to identify input features of the training data that are deemed the most important by the ML model for predicting positive outcomes. In contrast, \textit{example-based} explanations focus on identifying training data instances that are the most responsible for particular ML model decisions. Example-based explanations hinge on the \textit{valuation} of training data instances through popular data valuation techniques such as influence functions~\cite{koh2017understanding,icml2020_4261,cook-influence,10.1145/3514221.3517886} and data Shapley values~\cite{DBLP:conf/icml/GhorbaniZ19}. These explanation techniques have mostly explored ML model performance in terms of model accuracy or loss, but not for model fairness. Recently, Gopher~\cite{10.1145/3514221.3517886} generated example-based explanations for debugging instances of fairness violations for parametric ML models with twice-differentiable convex loss functions. None of these techniques are directly applicable to the problem of generating example-based explanations for fairness violations in tree-based models.

\noindent\textbf{Machine Unlearning.}
With the introduction of international regulations, such as the European Union’s General Data
Protection Regulation (GDPR)~\cite{gdpr}, the California Consumer Privacy Act (CCPA)~\cite{ccpa}, and  Canada’s proposed
Consumer Privacy Protection Act (CPPA)~\cite{cppa}, that center around users' \textit{right to be forgotten}, and 
concerns around the security and privacy of users' data, the field of \textit{machine unlearning}~\cite{10.5555/3454287.3454603,machine_unlearning_survey,ullah2021machine,gupta2021adaptive,bourtoule2019machine,pmlr-v139-brophy21a,schelter2021hedgecut,10.1145/3580305.3599420,10.1145/3548606.3559352,10.1145/3576915.3616585,10.1145/3583780.3615235,10.1145/3539618.3591989} has gained much popularity in recent years. 
The need for machine unlearning also stems from concerns around security and privacy of individuals whose data is present in the underlying training data for the ML model. Malicious actors might get access to individuals' sensitive data by exploiting system security vulnerabilities or by inference through model's predictions~\cite{ullah2021machine}; these dangers mandate that when an individual requests their data to be removed, it is not enough to remove it from the training data but also that their effect on the model be \textit{unlearned}~\cite{machine_unlearning_survey}.
%
Machine unlearning is in nascent stages; however, the idea of using unlearning to forget the effect of data instances on a model has shown promising results. Researchers have only recently started to explore the 
fairness implications of unlearning techniques~\cite{zhang2023forgotten}, and found that some frameworks (e.g., SISA~\cite{bourtoule2019machine}) have minimal impact on fairness for both uniform and non-uniform data deletions. We investigated the fairness impacts of DaRE-RF and observed that with both random and coherent data deletions, DaRE-RF mostly preserves fairness. However, this line of research needs further investigation. To the best of our knowledge, \sys is the first system that uses the concept of machine unlearning for the problem of fairness debugging. Although we focus on unlearning in random forest classifiers, the techniques presented are extensible to other ML models; we leave leveraging unlearning for debugging instances of fairness violations in other ML models for future work.

\noindent\textbf{Debugging Data-based Systems.}
Researchers have for long considered debugging training data as a way to explain the performance of data-driven systems, especially those using an underlying ML model~\cite{47966,10.1145/3448016.3457323,10.1145/3514221.3517864,10.1145/2723372.2750549,10.1145/3318464.3389763,10.1145/3514221.3517849,wu-complaint-driven}. Recently, Slice Finder~\cite{47966} and SliceLine~\cite{10.1145/3448016.3457323} identify \textit{slices} of the training data where the model performs worse compared to the rest of the data. Rain~\cite{wu-complaint-driven} determines parts of the training data responsible for an unexpected outcome in SQL queries involving ML decisions.
These efforts have focused on typical model performance metrics, such as accuracy and log loss, which are additive in nature. In contrast, the group fairness metrics we consider are not additive and hence, these techniques are not directly applicable to our problem setting.
\section{Conclusion}

This paper proposed \sys, a system for identifying the top-k coherent subsets responsible for instances of group fairness violations in the outcomes of a random forest classifier. \sys hinges on machine unlearning techniques to efficiently compute the contribution of subsets toward model bias, and utilizes a lattice-based search method based on the apriori algorithm in frequent itemset mining to prune the huge subset search space. To the best of our knowledge, the present work is the first to leverage machine unlearning in the context of fairness to explain the discriminatory behavior of a machine learning model. Through experimental evaluation on standard real-world datasets in the fairness literature, we demonstrate that the explanations generated by \sys identify subsets that are responsible for a substantial fraction of the model bias, and are consistent with prior studies on these datasets.

Our system focused on generating explanations for random forest classifiers; however, the approach described here can easily be applied to other (tree-based) classifiers. We consider further investigations on utilizing machine unlearning for fairness debugging of black-box ML models for future work.

\balance
\bibliographystyle{abbrv}
\bibliography{ref}
\end{document}